\documentclass[letterpaper, 10 pt, conference]{ieeeconf}  % Comment this line out if you need a4paper

\IEEEoverridecommandlockouts                              % This command is only needed if 
\usepackage{graphics} % for pdf, bitmapped graphics files
\usepackage{epsfig} % for postscript graphics files
\usepackage{mathptmx} % assumes new font selection scheme installed
\usepackage{cite}

\title{\LARGE \bf
How to select and use tools? : Active Perception of Target Objects Using Multimodal Deep Learning
}

\author{Namiko Saito$^{1}$, Tetsuya Ogata$^{2}$, Satoshi Funabashi$^{1}$, Hiroki Mori$^{3}$ and Shigeki Sugano$^{1}$% <-this % stops a space
\thanks{*This research was partially supported by JSPS Grant-in-Aid for Scientific Research (A) No. 19H01130, and the Research Institute for Science and Engineering of Waseda University.}% <-this % stops a space
\thanks{$^{1}$Namiko Saito, Satoshi Funabashi and Shigeki Sugano are with Department of Modern Mechanical Engineering, Waseda University, Tokyo, Japan {\tt\small n\_saito@sugano.mech.waseda.ac.jp, s\_funabashi@sugano.mech.waseda.ac.jp, sugano@waseda.jp}}%
\thanks{$^{2}$ Tetsuya Ogata is with the Department of Intermedia Art and Science, Waseda University, Tokyo, Japan, and the National Institute of Advanced Science and Technology, Tokyo, Japan {\tt\small ogata@waseda.jp}}%
\thanks{$^{3}$ Hiroki Mori is with the Future Robotics Organization, Waseda University, Tokyo, Japan {\tt\small mori@idr.ias.sci.waseda.ac.jp}}%
}

\begin{document}

\maketitle
\thispagestyle{empty}
\pagestyle{empty}

%%%%%%%%%%%%%%%%%%%%%%%%%%%%%%%%%%%%%%%%%%%%%%%%%%%%%%%%%%%%%%%%%%%%%%%%%%%%%%%%
\begin{abstract}

Selection of appropriate tools and use of them when performing daily tasks is a critical function for introducing robots for domestic applications. In previous studies, however, adaptability to target objects was limited, making it difficult to accordingly change tools and adjust actions. To manipulate various objects with tools, robots must both understand tool functions and recognize object characteristics to discern a tool--object--action relation. We focus on active perception using multimodal sensorimotor data while a robot interacts with objects, and allow the robot to recognize their extrinsic and intrinsic characteristics. We construct a deep neural networks (DNN) model that learns to recognize object characteristics, acquires tool--object--action relations, and generates motions for tool selection and handling. As an example tool-use situation, the robot performs an ingredients transfer task, using a turner or ladle to transfer an ingredient from a pot to a bowl. The results confirm that the robot recognizes object characteristics and servings even when the target ingredients are unknown. We also examine the contributions of images, force, and tactile data and show that learning a variety of multimodal information results in rich perception for tool use.

\end{abstract}

%%%%%%%%%%%%%%%%%%%%%%%%%%%%%%%%%%%%%%%%%%%%%%%%%%%%%%%%%%%%%%%%%%%%%%%%%%%%%%%%
\section{INTRODUCTION}

%House helping robot and the benefit of tool-use
A tool is an interface for manipulating an object, extending the user’s manipulation ability in terms of reachable space, resolution, and the direction and strength of force. Daily tasks in a complicated environment require various actions, and many tasks cannot be solved by hands alone. By selecting and handling tools, however, humans and even robots can adaptively achieve such tasks. We aim at allowing robots to handle daily tools designed for humans.

%Adaptability to objects
Adaptability to various target objects and operations is a central issue for tool-use by robots. Most robots are designed to repeat specific tool-use motions for specific objects \cite{Pan, Bread, Spatula, Beetz}. Some research has realized robots that can generate or choose suitable actions depending on the available tools \cite{Stoytchev, Takahashi, Nabeshima, Mar, Fang}. These studies succeeded in tool manipulation, but it was difficult to change actions according to the target objects, and when required to handle a new object, the control systems need to be redesigned. To resolve this issue, robots need to discern tool--object--action relations. That is, they must not only understand tool functionality, but also be able to determine which tool should be used and adjust how they handle it depending on the object. This idea has been discussed as affordance theory \cite{Gibson, Jamone, Zech}.

%Recognition of extrinsic and Intrinsic characteristics
Recognition of extrinsic and intrinsic characteristics is critical when dealing with various objects. Extrinsic characteristics are related to appearance, such as shape, color, size, posture, and position. Intrinsic characteristics include weight, friction, and hardness. Extrinsic characteristics are thus visible, while intrinsic ones are perceived by tactile and proprioceptive senses. Many factors for deciding actions depend on both characteristics, such as which tools to use, suitable movement speed, approach trajectory, and strength for grasping a tool. If robots cannot recognize intrinsic characteristics in particular, they may fail to move heavy objects, knock over soft objects, or allow low-friction objects to slip from their grasp. Moreover, objects may be similar in appearance but have different intrinsic characteristics, or vice versa, so both characteristics should be recognized accurately. Some previous studies have considered the extrinsic characteristics of objects and their manipulation with tools, but did not consider intrinsic characteristics \cite{Dehban, Saito-Humanoid, Xie}. To the best of our knowledge, this study is the first to consider both types of characteristics during tool use.

%Active perception, multimodal learning
We focus on ``active perception'' to realize recognition of both extrinsic and intrinsic characteristics. Active perception allows recognition of target characteristics by capturing time-series changes in sensory information while directly acting on and interacting with a target \cite{Active, Active-manipulation, Benjamin}. We allow a robot to first interact with an object for a while to recognize its extrinsic and intrinsic characteristics, assuming that a variety of multimodal information lets the robot better recognize those characteristics.% We examine the contributions of images, force, and tactile information and show the effect of multimodal learning on this task.

%Why we choose serving food: Tool-selection and handle it
We use transferring food ingredients as an example tool-use task because this is one of the most common situations requiring use of varied tools depending on the target objects. This task requires the robot to select a proper tool and adjust its handling depending on the extrinsic and intrinsic characteristics of the ingredients. Preparing several ingredients in a pot, we allow the robot to recognize what ingredients are inside, select a ladle or turner depending on the ingredient characteristics, and transfer the ingredient to a bowl.

%summary
We construct a DNN model that allows the robot to recognize the characteristics of the target ingredient by active perception while stirring the ingredient for a while, then select a turner or a ladle and handle it accordingly. The DNN model comprises two modules: a convolutional autoencoder (CAE) \cite{CAE} and a multiple timescales recurrent neural network (MTRNN) \cite{MTRNN}. To discern the tool--object--action relation, recognize object characteristics, and generate appropriate tool-use actions, we apply the CAE to compress raw images and the MTRNN to predict and generate multimodal sensorimotor time-sequence data. We test the DNN model on a humanoid robot, Nextage Open developed by Kawada Robotics. For evaluation, we use untrained ingredients and confirm whether the robot can select and use proper tools for handling them. We also analyze the role of image, force, and tactile information in task execution by comparing success rates while varying combinations of input sensorimotor data. Our contributions are as follows. 
\begin{itemize}
    %\item The DNN model does not require a pre-designed model of the target objects, only to learn sensorimotor data for dealing unknown objects.
    \item The developed robot can conduct tool-use depending on tool--object--action relations, that is, select a tool and manipulate it depending on the target object.
    \item The developed robot can recognize extrinsic and intrinsic characteristics of even unknown objects by active perception.
    \item The DNN model allows the robot to seamlessly pursue whole tool-use tasks, recognizing object characteristics and selecting and picking a proper tool for handling it.
    %\item We reveal contributions of each sensor information for the tool-use task.
\end{itemize}

\section{Related Works}

\subsection{Tool-use by robots}

%tool funciton
Some related works on tool-use by robots realized robots capable of recognizing the characteristics of tools and handling them. Stoytchev and Takahashi et~al.~\cite{Stoytchev, Takahashi} developed a robot capable of learning differences in tool functions depending on their shape. These studies allowed each robot to manipulate tools according to their shape and to move an object to a goal. Nabeshima et~al. developed a robot capable of manipulating tools to pull an object from an invisible area, regardless of the shape of the grasped tool~\cite{Nabeshima}. Mar et~al. considered the tool’s attitude, allowing a robot to know in which direction to push to efficiently move an object~\cite{Mar}. In addition, Fang et~al \cite{Fang}. realized a robot to consider grasping direction and point of tools as well. In these studies, experimenters passed the robots one tool, rather than allowing them to select a tool themselves, and their target objects were specified. In other words, those studies did not consider robots capable of discerning tool--object--action relations, selecting tools, or adjusting their movements depending on object characteristics.

% Tool Selection
Some studies have tackled discernment of tool--object--action relations and realized tool selection and manipulation depending on target objects. Dehban et~al. constructed Bayesian networks that express relations among tools, objects, actions, and effects, allowing a robot to predict or select one factor by the other three factors~\cite{Dehban}. However, the actions were predetermined, with the robot following designated motions, so it was difficult to flexibly adapt to uncertain situations, especially object behaviors such as rolling, falling, and sliding in different ways. On the other hand, Saito et~al. developed a robot capable of selecting one of two tools placed in front of it depending on the initial and target object positions in \cite{Saito-Humanoid}. Moreover, Xie et al. realized a robot to pick up and flexibly use tools to move objects to ordered places \cite{Xie}. However, these studies considered only the appearance of the objects, not their intrinsic characteristics. 

\subsection{Recognition of object characteristics}

%active perception refer
Some research has focused not on tool-use, but on acquiring extrinsic and intrinsic characteristics of target objects, and we referred to these studies in our efforts to realize a robot capable of recognizing both types of characteristic through active perception. Balatti et~al. realized a robot capable of scooping several kinds of soil~\cite{iit-dig}. The robot discerned soil characteristics like moisture and weight by stirring the soil with a stick, then used image and force sensor data to calculate appropriate arm trajectories for scooping according to those characteristics. Saito et~al. and Lopez-Guevara et~al. realized a robot capable of pouring liquid from a container~\cite{Saito-ROBIO, Lopez}, and allowed the robot to determine the features of the liquid like amount and viscosity from sensor data while shaking or stirring the container. Schenck et~al. realized a robot to dump a granular object to deform into a desired shape\cite{Schenck}. In their study, the robot recognize the dynamics of the object while scoops and make suitable action for dumping. In summary, these studies first allowed the robot to interact with the target objects to perceive their characteristics, then allowed the robot to pursue its task according to that perception. We apply these ideas of active perception to the tool-use tasks in this study.

\section{Method}

\subsection{Tool-use task}
The robot task comprises an ``active perception part'' and a ``motion generation part.'' During the active perception part, the robot stirs the ingredients in the pot. The stirring motion is the same every time, but object behaviors and input sensor data are different depending on the ingredients, and the DNN model uses those differences to recognize intrinsic and extrinsic characteristics of the objects. Before we start experiments, we tried controlling the robot to perform several patterns of stirring manually, and adopted the stirring way in which the robot could interact with objects most, in other words, different behavior of sensor information depending on objects is most likely to appear. The motion is to go back and forth from the right end to the left end of the pot with a wooden spatula.

During the motion generation part, the robot acts considering tool--object--action relations. Specifically, the robot selects a suitable tool, grasps it with appropriate strength, and adjusts its arm joint angles to determine an appropriate trajectory for transferring the ingredients. Appropriate motions will differ every time even if the ingredient is the same, because the ingredient’s position and attitude is varied. The robot does not use fixed motions but generates motions step by step in real time for picking up a tool, holding an ingredient and transferring it to a bowl. If the action does not match to the situation, the robot will fail at the task.

\subsection{Experiment}

\begin{figure}
\centerline{\includegraphics[width=8.5cm]{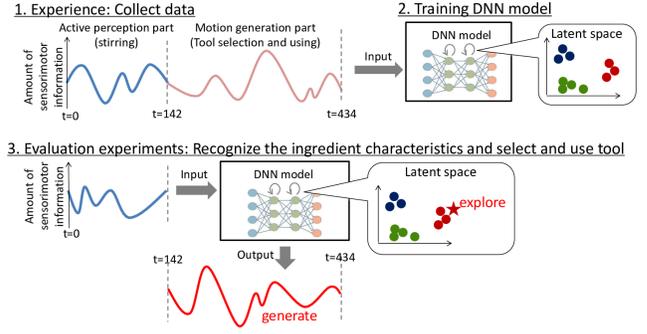}}
\caption{Overview of the method. The task comprises the active perception and the motion generation parts. We first manually control the robot to collect training data. Second, we train the DNN model and the model learns to express the characteristics of ingredients and tool--object--action relations in the latent space. Finally, the robot is tested to transfer ingredients by itself. After performing active perception, the DNN model explores the latent space to generate motions for selecting and picking up a tool and transferring an ingredient.}
\label{method}
\end{figure}

Figure~\ref{method} shows the overall process. We first allow the robot to experience several transfer tasks with some ingredients to accumulate training data by manually controlling. The robot performs both active perception (fixed stirring motion) and motion generation (selecting a tool and transferring an ingredient, which is remotely controlled by experimenters). We record sensorimotor data during that time, then use this data to train the DNN model. The DNN model thereby learns to recognize ingredient characteristics and acquire tool--object--action relations in the latent space, which resulting in determining tool selection and the way to manipulate it.

We evaluate two experiments: Generalization evaluation and Evaluation on contribution of multimodality. In that evaluations, experimenters simply put an ingredient in a pot, and the robot itself must detect its characteristics and generate appropriate motions. While the robot stirs (active perception), the DNN model explore the values in the latent space that best describe the ingredient's characteristics and tool handling information, and the robot generates motions accordingly.

\subsubsection{Generalization evaluation}

The robot transfers several untrained ingredients to confirm the generalization ability of the DNN model. We conduct principle component analysis (PCA) on the latent space to evaluate whether the robot can recognize ingredient characteristics, suitable tool to use, and the way to use the tool. We perform the tests five times each and confirm whether the robot succeed in four task stages: tool selection, tool grasping, picking up the ingredient, and pouring the ingredient.

\subsubsection{Evaluation on contribution of multimodality}

We investigate the contribution of information from each sensor by varying the input data to the DNN models in the combinations shown below. As in Generalization evaluation, we compare success rates at four task stages. We perform tests with training ingredients sixty times each (6 kind of ingredients~$\times$ 2 amounts~$\times$ 5 times). Analyzing the contribution of various sensor data allows us to confirm the effect of multimodal learning for tool-use.
\begin{itemize}
    \item Image feature data + force sensor data + tactile sensor data (proposed method)
    \item Image features + force
    \item Image features + tactile
    \item Image features
    \item Force + tactile
\end{itemize}

%
%%%%%%%%%%%%%%%%%%%%%%%%%%%%%%%%%%%%%%%%%%%%%%%%%%%%%%%%%%%%%%%%%%%%
\section{Deep Learning Model}

\begin{figure}
\centerline{\includegraphics[width=8cm]{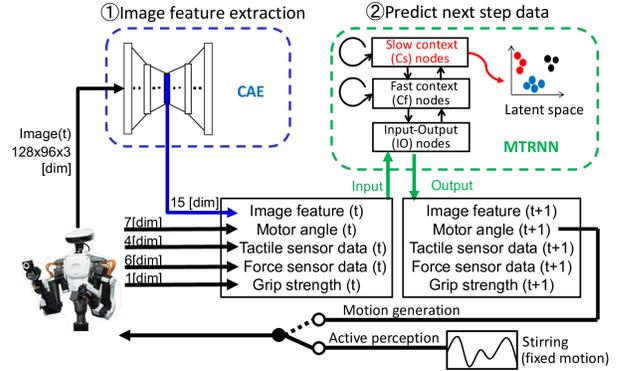}}
\caption{Overview of the DNN model, which consists of a CAE that extracts image features and an MTRNN that generates next-step motions.}
\label{model}
\end{figure}
Figure~\ref{model} shows an overview of the DNN model, which consists of two components: a CAE and an MTRNN. This model is taken from \cite{Saito-Humanoid}, however we change the way to utilize it. In \cite{Saito-Humanoid}, they use the DNN for only forward calculation. In the current paper, we firstly apply the DNN to active perception for exploring the latent space of the DNN model, resulting in recognition of an object's intrinsic and extrinsic characteristics. Then the same DNN is used to generate motions with the explored value.

%%%%%%%%%%%%%%%%%%%%%%
\subsection{Image feature extraction by the CAE}

\begin{table*}[h!]
\caption{Structure of the CAE}
\begin{center}
\begin{tabular}{lcccccc}
\hline\hline
\textbf{The /th Layer} & \textbf{Input}& \textbf{Output} & \textbf{Processing}& \textbf{Kernel Size} & \textbf{Stride} & \textbf{Padding}  \\
\hline
1&(128, 96, 3)&(64, 48, 32)&convolution&(4,4)&(2,2)&(1,1)\\
2&(64, 48, 32)&(32, 24, 64)&convolution&(4,4)&(2,2)&(1,1)\\
3&(32, 24, 64)&(16, 12, 128)&convolution&(4,4)&(2,2)&(1,1)\\
4&(16, 12, 128)&(8, 6, 256)&convolution&(4,4)&(2,2)&(1,1)\\
5&(8, 6, 256)&(4, 3, 512)&convolution&(4,4)&(2,2)&(1,1)\\
6&6144&254&fully connected&\textbf{-}&\textbf{-}&\textbf{-}\\
7&254&15&fully connected&\textbf{-}&\textbf{-}&\textbf{-}\\
8&15&254&fully connected&\textbf{-}&\textbf{-}&\textbf{-}\\
9&254&6144&fully connected&\textbf{-}&\textbf{-}&\textbf{-}\\
10&(4, 3, 512)&(8, 6, 256)&deconvolution&(4,4)&(2,2)&(1,1)\\
11&(8, 6, 256)&(16, 12, 128)&deconvolution&(4,4)&(2,2)&(1,1)\\
12&(16, 12, 128)&(32, 24, 64)&deconvolution&(4,4)&(2,2)&(1,1)\\
13&(32, 24, 64)&(64, 48, 32)&deconvolution&(4,4)&(2,2)&(1,1)\\
14&(64, 48, 32)&(128, 96, 3)&deconvolution&(4,4)&(2,2)&(1,1)\\
\hline
\end{tabular}
\end{center}
\label{CAE_model}
\end{table*}

A CAE has an hourglass structure that learns to make output data the same as input data. The loss function is as
\begin{equation}
E=0.5\sum\left(y_i-x_i\right)^2,
\end{equation}
where $i$ is all the data in all the steps. This allows compression of raw images, with image features extracted from an intermediate (7th) layer, whose number of dimensions is the lowest. Table I shows construction of the CAE. We use ReLU as the activation function except in the center layer, where we use the Sigmoid function because its output range is suitable for the next input to the MTRNN. Network weights are updated using stochastic gradient descent. Raw camera images initially have 36,864 dimensions (128 (width)~$\times$ 96 (height)~$\times$ 3 (channels)), but that data is compressed into 15-dimensional image features. We trained the CAE over 1,000 epochs.

%%%%%%%%%%%%%%%%%%%%%%%%%
\subsection{Motion generation by the MTRNN}

\begin{table}[]
	\begin{center}
		\caption{MTRNN node setting}
		\begin{tabular}{c|c|c}\hline \hline
			Nodes & Time constant & Number of nodes \\ \hline
			${\rm {C_f}}$ & 70 & 10 \\ \hline
			${\rm {C_f}}$ & 5 & 100 \\ \hline
			IO & 2 & 33 (Image feature~(15)+Motor~(7)\\ 
			 &  &  +Tactile~(4)+Force~(6)+Grip~(1))\\ \hline
		\end{tabular}	
		\label{mtrnn_tau}
		%
		%\caption{Setting of feedback rate $\alpha$}
		%\begin{tabular}{c|c|c}\hline \hline
		%	Situation & Part & Value of $\alpha$ \\ \hline
		%	Training & Both parts & 0.9 \\ \hline
		%	Evaluation & Active perception part & 1.0 \\ \cline{2-3}
		%	Experiments& Motion generation part & 0.8 \\ \hline
		%\end{tabular}
		%\label{alpha}	
	\end{center}
\end{table}
An MTRNN is a recurrent neural network that predicts a next step from current state data. An MTRNN comprises three types of nodes with different time constants: slow context (${\rm {C_s}}$) nodes, fast context (${\rm {C_f}}$) nodes, and input/output (IO) nodes. Because of their large time constant, ${\rm {C_s}}$ nodes learn data sequences, whereas ${\rm {C_f}}$ nodes with a small time constant learn detailed motion primitives. This enables learning of long and dynamic time-series data. 

Table I\hspace{-.1em}I shows settings of each node. We tried several combinations and adopted the one that minimized training error most. If these numbers are too small, complex information cannot be learned, and if they are too large, the model is overtrained and cannot adapt to untrained data.
%Table I\hspace{-.1em}I shows settings for the time constants and numbers of each node. We varied numbers of ${\rm {C_s}}$ nodes in the range of 8 to 12, the time constant of ${\rm {C_s}}$ nodes in the range of 50 to 100, and numbers of ${\rm {C_f}}$ nodes in the range of 50 to 150 in increments of 10. We adopted the combination that minimized training error. If these numbers are too small, complex information cannot be learned, and if they are too large, the model is overtrained and cannot adapt to untrained data. The time constant of ${\rm {C_f}}$ had little effect, even when it was changed to around 5.

\subsubsection{Forward calculation}

Output values from MTRNN are calculated as follows. First, the internal value \(u_i\) of the neuron \(i\) at step \(t\) is calculated as
\begin{equation}
u_i(t)=\biggl(1-\frac{1}{\tau_i}\biggr)u_i(t-1)+\frac{1}{\tau_i}\left[\sum_{j\in {\rm {N}}}w_{ij}x_j(t)\right],
\end{equation}
where \({\rm {N}}\) is the number of neurons connected to neuron \(i\), \(\tau_i\) is the time constant of neuron \(i\), \(w_{ij}\) is the weight value from neuron \(j\) to neuron \(i\), and \(x_j(t)\) is the input value of neuron \(i\) from neuron \(j\). Then the output value is calculated using the hyperbolic tangent function
\begin{equation}
y_i(t) =  {\rm {tanh}}\left(u_i(t)\right).
\end{equation}
as the activation function. The value of \(y_i(t)\) is used as the next input value as
\begin{equation}
x_i(t+1)=\left\{ \begin{array}{ll}
\alpha\times y_i(t)+(1-\alpha)\times {\rm {T}}_i(t+1) & i\in{{\rm {IO}}}\\\\
y_i(t) & {\rm {otherwise}}
\end{array}\right.
\label{eq:x_i},
\end{equation}
where \({\rm {T}}_i(t)\) is the input datum \(i\), which is training data during model training or real-time data during evaluation experiments. If neuron \(i\) is an IO node, the input value \(x_i (t)\) is calculated by multiplying the output of the preceding step \(y_i(t-1)\) and the datum \({\rm {T}}_i(t)\) by the feedback rate \(\alpha\) (\(0 \le \alpha \le 1\)). We can adjust the input data by means of the feedback rate \(\alpha\). If the value is large, the model can stably predict the next step, whereas if the value is small, the model can easily adapt to real-time situations. It is set 0.9 during training the DNN model, wheres during evaluation, it is set 1.0 in the active perception part and 0.8 in the motion generation part. We tried setting $\alpha$ testing next-step data predictions using test data taken in advance. Finally, we adopted $\alpha$ values that gave the best result and allowed adaption to individual circumstances.
%During training, the model can be trained efficiently by setting $\alpha$ to 0.9, that is, combining the predicted data 90 \% and training data 10 \%. On the other hand when we do evaluation experiment, we set the value to 0.8, that is, combining the predicted data 80 \% and real time data 20 \%. We tried setting $\alpha$ to 0.8, 0.9, and 1.0, and finally adopted each value which gave the best result and make it possible to adapt to each individual circumstances.  

\subsubsection{Backward calculation during training}

We use back propagation through time (BPTT) algorithm \cite{BPTT} to minimize the training error given by
\begin{equation}
E=\sum_{i}\sum_{i\in{{\rm {IO}}}}\left(y_i(t-1)-{\rm {T}}_i(t)\right)^2
\label{y}
\end{equation}
and update the weight as
\begin{equation}
w^{n+1}_{ij}=w^{n}_{ij}-\eta \frac{\partial E}{\partial w^{n}_{ij}}
\label{weight}
\end{equation}
where \(\eta(=0.0001)\) is the learning rate and \(n\) is the number of epochs.

Simultaneously, to make the initial step value of Cs layer (Cs(0)) perform as the latent space, we update it as
\begin{equation}
{\rm {C_s}}^{n+1}_i(0)={\rm {C_s}}^n_i(0)-\eta \frac{\partial E}{\partial {\rm {C_s}}^n_i(0)}. \label{Cs}
\end{equation}
The model training starts with all ${\rm {C_s}}(0)$ values set to 0. As a result, ${\rm {C_s}}(0)$ space stores the features of the dynamics. We expect that information regarding ingredient characteristics and tool--object--action relations self-organize in it. 

\subsubsection{Exploring latent space value by active perception during evaluation}

In evaluation experiments, the DNN model needs to explore the latent ${\rm {C_s}}(0)$ values which match to the ingredient since by inputting proper ${\rm {C_s}}(0)$ values to the trained MTRNN, the DNN model can generate proper motions. The latent ${\rm {C_s}}(0)$ values is explored by Eq. (\ref{Cs}) with BPTT during the active perception part by setting the error as 
\begin{equation}
E=
\sum_{t=1}^{{\rm {t_{ap}}}}\sum_{i\in{{\rm {sensor}}}}\left(y_i(t-1)-{\rm {T}}_i(t)\right)^2, 
\label{recog}
\end{equation}
where \({\rm {t_{ap}}}\) is the final step of the active perception part and datum \({\rm {T}}_i(t)\) is real-time sensorimotor data. We only use the sensor data (image features, tactile data, and force data) because the motor data for stirring is every time the same. The MTRNN predicts sensor data while the robot stirring and calculates the error between the prediction and the actual input. Since the feedback rate \(\alpha\) in Eq. (\ref{eq:x_i}) is set 1.0 in the active perception part, the output $y_i$ is predicted just by ${\rm {C_s}} (0)$ and $x(0)$. Moreover, we do not update the weight values with Eq. (\ref{weight}). Therefore, what the model can do to minimize the error is only adjusting ${\rm {C_s}} (0)$. The calculations is performed over 10,000 epochs when the error was already fully converged.

The exploring calculation ends in short time, thus the DNN model immediately and seamlessly shift to the motion generation part. In the motion generation part, the DNN model predicts the next-step data with forward calculation using the explored ${\rm {C_s}}(0)$ values. The robot moves according to the predicted motor angles and grip strengths to grasp a tool and transfer an ingredient. The method of exploring the latent space and utilizing it for robot control is the novelty of this study.

%%%%%%%%%%%%%%%%%%%%%%%%%%%%%%%%
\section{Experimental Setup}
\subsection{System design}

We use a humanoid robot (Nextage Open; Kawada Robotics \cite{Nextage}), controlling it to perform tasks with its right arm, which has six degrees of freedom. The robot has two head-mounted cameras, but we use only the right eye camera, which produces images with 36,864 dimensions (height 128 $\times$ width 96 $\times$ 3 channels). We attach a force sensor (DynPick; Wacoh-Tech) in its wrist and record six-axis force data. In addition, we attach a gripper (Sake Hand; Sake Robotics). We also attach a tactile sensor (POTU-001-1; Touchence Inc.) with four sensing points on the surface of its gripper.

We record robot arm joint and grip angle data (7~dim), grip strength (1~dim), and camera images every 0.1 sec. We record tactile data (4~dim) and force data (6~dim) every 0.025~sec, averaging sensor values over four times and sampling every four times. The sampling frequency of all the sensorimotor data is then 0.1~sec, or 10~Hz.

Arm and gripper angle data, grip strength, tactile data, and force data input to the MTRNN are normalized to $[-0.9, 0.9]$ before inputting to the model. Image data is input to the CAE at its original range, $[0, 255]$.

We use a total of 48 time-series datasets as training data (6 ingredient types~$\times$ 2 amounts $\times$ 4 times).

%%%%%%%
\subsection{Tools and objects}

\begin{figure}[]
		\begin{center}
			\includegraphics[width=7cm]{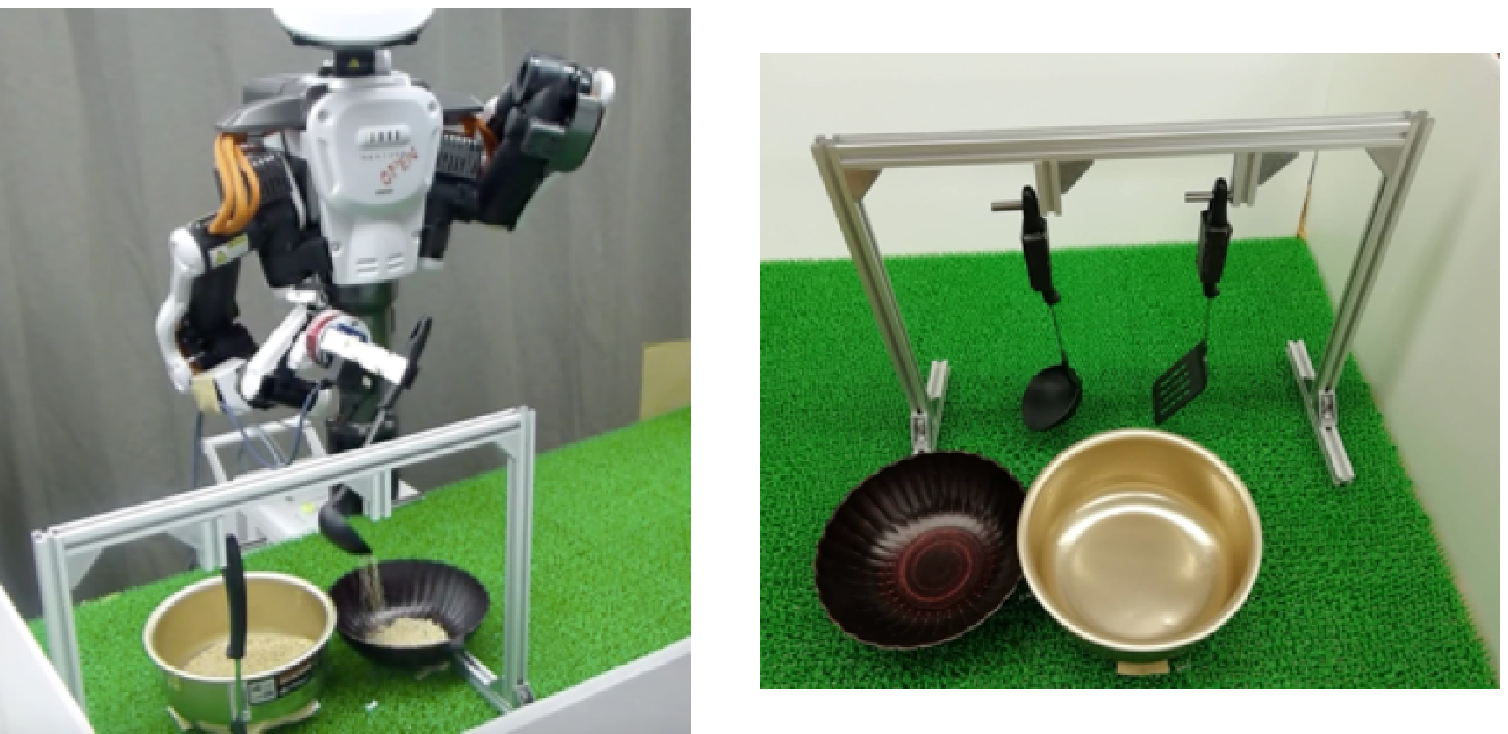}
			\caption{Table setting.}
			\label{setting}
		\end{center}
\end{figure}

We put a table in front of the robot and place a pot, a bowl, and a tool rack on the table (Fig.~\ref{setting}). We prepare a ladle and a turner as selectable tools for transfer, hung from the rack at the same position every time. We also prepare a wooden spatula for stirring ingredients during the active perception part.

\begin{figure*}
\centerline{\includegraphics[width=15cm]{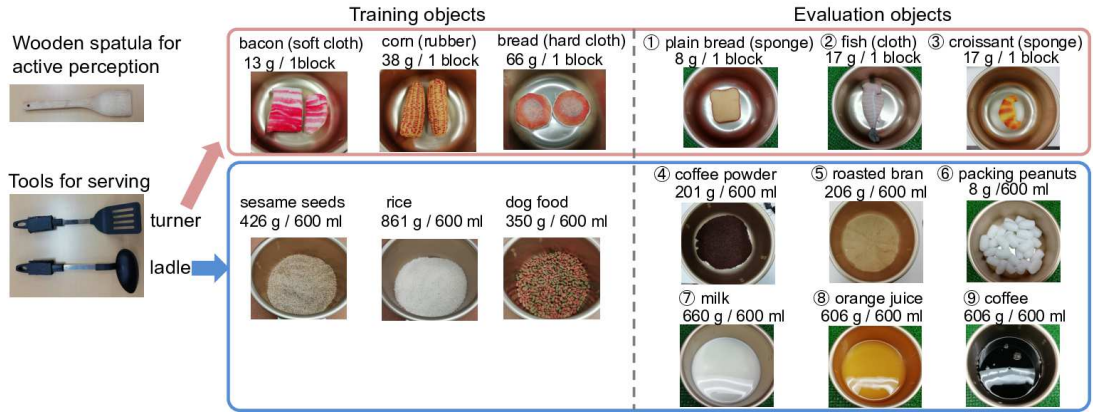}}
\caption{Tools and objects used in the experiments.}
\label{object}
\end{figure*}

Figure~\ref{object} shows the tools used, along with six of the ingredients used for training. We prepared the ingredients so that the DNN model can be trained with variety of objects. We prepare three granular ingredients (sesame seeds, rice, and dog food) best handled with a ladle. We perform training using 300 ml and 600 ml of these ingredients. We prepare three blocklike simulated ingredients (bacon made from soft cloth, rubber corn kernels, and bread made from hard cloth) best handled using a turner. We train using one and two of these blocks. Whereas we use nine kinds of untrained ingredients (three each of granular, liquid, and block ingredients) for evaluation. We prepare 600 ml or 1 block of each.


\subsection{Task design}

The robot starts every motion from a specific home position, initially holding the wooden spatula. As the active perception part, the robot stirs an ingredient in the pot with the wooden spatula for 14.2~sec. Experimenters then control the robot to open its gripper and take the wooden spatula from its hand. We do not record sensorimotor data during this gap time. The model then seamlessly shifts to the motion generation part. The robot generates each motion for selecting and grasping a tool, manipulating the ingredient with a tool to transfer it to a bowl. The robot then comes back to the home position. The time length is 29.2~sec. The total time is thus 43.4~sec, and there are 434 steps.

The inclination at which the tool should be held must be changed depending on the manipulated ingredient. In addition, if the ingredient is heavy or has high friction, the robot must grasp the tool more tightly. For power efficiency, the robot should use as little gripping force as possible. If the strength and inclination are not suited to the ingredient, the robot will fail at transferring it. 

We define a successful trial as the robot transferring more than 1 block or 50 ml of the ingredient to the bowl without dropping the tool or spilling the ingredient.

%%%%%%%%%%%%%%%%%%%%%%%%%%%%%%%%%%%%%%
\begin{figure}
\centerline{\includegraphics[width=8cm]{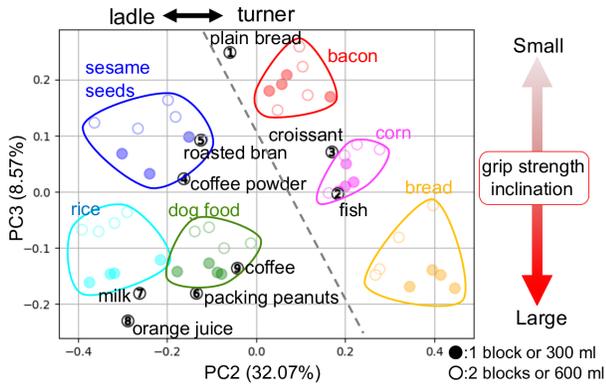}}
\caption{Principal component analysis on the latent space, ${\rm {C_s}}(0)$. The results indicate recognition of object characteristics and acquisition of tool--object--action relations. We are also showing the contribution rate of PC2 and PC3. The PC1 axis, which is not shown in the figure related to ingredient color.}
\label{PCA}
\end{figure}

\begin{figure}
\centerline{\includegraphics[width=8cm]{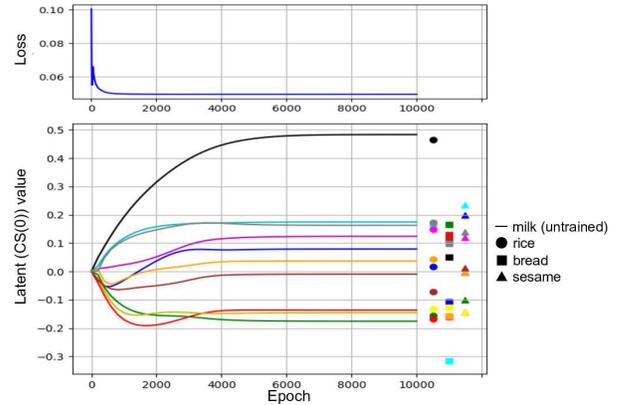}}
\caption{Exploring ${\rm {C_s}}(0)$ values of milk during active perception. The circle, square and triangle plots indicate 600 ml of rice, 1 block of bread and 600 ml of sesame, each. The explored values are most similar to ones of rice}
\label{CsExplore}
\end{figure}
\section{Results and Discussion}

\subsection{Generalization evaluation}

Figure~\ref{PCA} shows the results applying PCA to the latent space, with ${\rm {C_s}}(0)$ values for transferring training ingredients plotted as circles. In the figure, plot colors indicate the ingredients, and fillings indicate their amount. Plots of similar ingredients are clustered, and their placement shows the tool and handling method best suited to them. The PC2 axis whose contribution rate is 32.07\% shows the appropriate tool and the PC3 axis whose contribution rate is 8.57\% shows action information, namely the grasping strength and holding inclination for the tool suited to the ingredient. This shows that the DNN model could self-organize ingredient characteristics and acquire tool--object--action relations in the latent space. Incidentally, the PC1 axis (not shown in the figure) related to ingredient color like the order as light to dark: rice and sesame (white), corn (yellow), bacon and bread (red), to dog food (brown). 

The explored ${\rm {C_s}}(0)$ values after stirring untrained ingredients are also mounted on Figure~\ref{PCA} with circled numbers. We can presume recognition of ingredient characteristics by the DNN model from plot placements. For example, milk plotted as the circle numbered 7 is near the region of rice, so we assume the DNN predicts that milk has characteristics similar to rice, and that it is proper to lift it using a ladle with a large grip strength and tilted at a large inclination. Milk and rice are both smooth and heavy liquid or granular, so their characteristics are similar. Other untrained ingredients are also plotted on or near regions with characteristics similar to the training ingredients. 

Figure~\ref{CsExplore} shows the loss and 10 dimensional ${\rm {C_s}}(0)$ values of milk explored with 10,000 epoch calculation during active perception. The circle, square and triangle plots indicate 600 ml of rice, 1 block of bread and 600 ml of sesame, each. The explored values are converged and resulting in similar to ones of rice. We thus conclude that the DNN model could recognize ingredient characteristics by active perception. We also suggest that the DNN model could use the latent space to determine which tools to be used and how to act. 

\begin{figure*}
\centerline{\includegraphics[width=15cm]{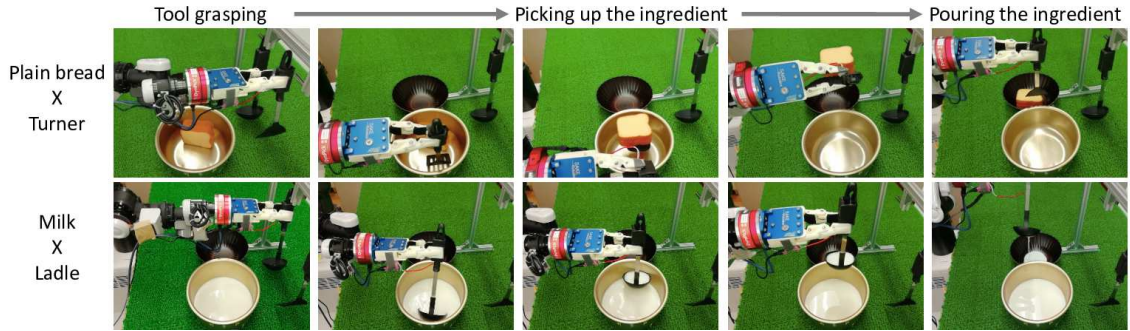}}
\caption{The robot generates motions by the DNN model and selects tools appropriate to untrained ingredients.}
\label{online}
\end{figure*}

\begin{figure*}
\centerline{\includegraphics[width=15cm]{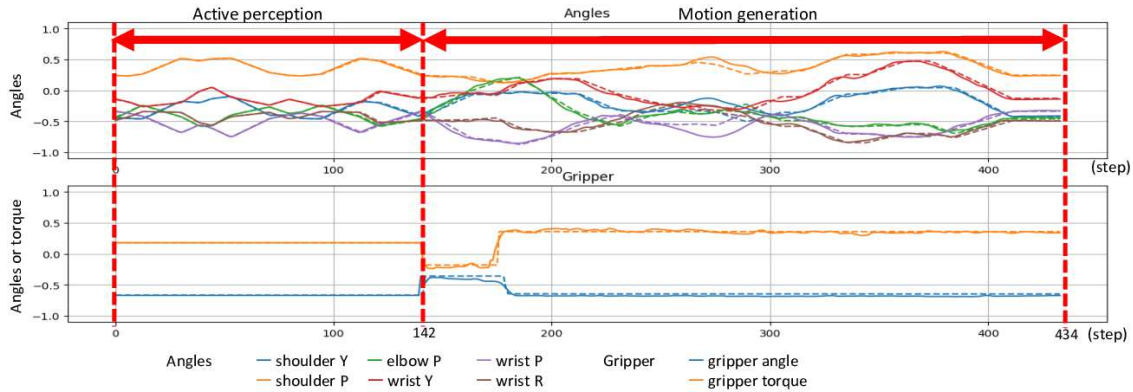}}
\caption{The whole trajectories of arm joint and gripper angles and gripper torque generated by the DNN model. All data are normalized to $[-0.9, 0.9]$. Solid lines show data from the robot transferring milk, an untrained ingredient. Dotted lines show data when transferring rice, a trained ingredient.}
\label{online_angle}
\end{figure*}

Figure~\ref{online} shows how the robot generated serving motions with untrained plain bread and milk. When transferring the plain bread, the robot selected a turner, adjusted its arm angle to tilt the turner at a small inclination to lift the bread, and transferred it to the bowl. When the robot transferred milk, it selected a ladle, adjusted its arm angle to tilt the ladle deeply to scoop up the milk, and transferred it to the bowl. The robot could thus select a tool and manipulate it according to ingredient characteristics.

Figure~\ref{online_angle} shows an example of arm trajectories, gripper angles, and grip strength during the whole task. All data are normalized to $[-0.9, 0.9]$. When the robot grasps a tool, it is necessary to adjust the direction, and perform the picking at an appropriate timing. Then, it is necessary to adjust the operating angle and the gripping strength according to not only ingredient characteristics but also the gripping position and pose of the tool. Furthermore, the manipulation needs to correspond to the ingredient's real time behavior. The figure shows generated data for milk (untrained) as solid lines compared with that for rice (trained) as dotted lines. The solid lines are close to the training data, but sometimes shifted during motion generation. The figure shows that the motor angles and grip strength were flexibly adjusted depending on the situation.

\begin{figure}
\centerline{\includegraphics[width=8cm]{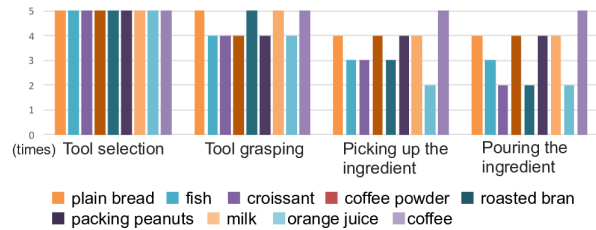}}
\caption{Success rate of experiment~1 with untrained ingredients.}
\label{online_result}
\end{figure}

We conducted these experiments using other untrained ingredients, too. Figure~\ref{online_result} summarizes the results. The robot succeeded in tool selection in all experiments and in grasping them in most, so the robot could select a suitable tool with high accuracy. As for lifting and transferring ingredients, there were more than two out of five successes for each. Success rates were especially poor for croissants, orange juice, and roasted bran. All the three ingredients are a similar color to the pot, so it may have been more difficult to recognize positions than in other trials.

\subsection{Evaluation on contribution of multimodality}

\begin{figure}
\centerline{\includegraphics[width=8.5cm]{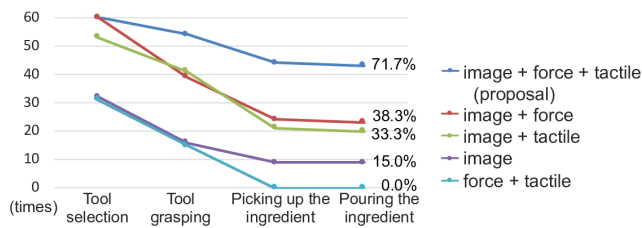}}
\caption{Success rates in experiment~2, which used trained ingredients and varied combinations of input sensors.}
\label{ex2}
\end{figure}

We used trained ingredients to investigate the contribution of sensor data by varying input data. Figure~\ref{ex2} summarizes the results, showing that the success rate of the proposed input (image + force + tactile sensor data) was the highest at each of the four stages: tool selection, tool grasping, picking up an ingredient, and pouring an ingredient. The success rate was 71.7\%, which was more than double the success rate of other inputs.

When we input only ``image'' or ``force + tactile sensor data,'' the success rate for tool selection is low, with nearly half of the trials failing. These results suggest that images and either force or touch are necessary at a minimum for active perception.

After tool selection, we compare rates for successful tool grasping (slopes of lines in the figure). ``Image + tactile'' has a higher success rate than ``image + force,'' so we can say that tactile sensor data contributes to grasping. During the ``image + force'' experiment, we observed many failures in which the robot moved near the target tool, but its grasp was slightly offset or the tool slipped.

As for the success rate for picking up and pouring ingredients, ``force + tactile'' failed in all trials, suggesting the necessity of using image data to perceive the extrinsic characteristics of ingredients and to handle the tool. ``Image + force'' has a higher success rate than ``image + tactile,'' suggesting that force sensor data contributes more toward perceiving intrinsic characteristics and handling tools than tactile data.

In summary, we found that image data was necessary to perceive extrinsic characteristics, and that force data was mainly used to perceive intrinsic characteristics. Tactile data was also used for perceiving intrinsic characteristics, but played a larger role in grasping tools. All sensor data were thus significant for active perception and task performance, and multimodality realized rich perception for adaptive tool use with a variety of objects.

\section{Conclusion}

In this paper, we introduced a general framework based on DNN model that acquires  tool–object–action relations which modifies actions of a robot through to the objective.
Active perception enabled the robot to recognize a target objects' characteristics and resulted in selecting and manipulating an appropriate tool. As a result, the robot succeeded in transferring untrained ingredients, conﬁrming that the DNN model could recognize extrinsic and intrinsic characteristics of unknown objects in its latent space. In addition, contributions of all sensor data were investigated and found that a variety of multimodality realizes rich perception for more successful tool-use manipulation. 

%impact and future work
This framework is a general approach for learning robotic tool-use and thus it could also be applied to other tool-use problems and robotic platforms. However, this study is limited to select one tool from two specific ones. In future works, we will realize that robots use a variety of tools by considering characteristics of the tools. Another limitation of this study is that there is a limit to handling ingredients that have significantly different characteristics from the trained ones. We will update the way to conduct active perception which can pull out much more difference in sensor information and enable more variety of characteristics.
%%%%%%%%%%%%%%%%%%%%%%%%%%%%%%%%%%%%%%%%%%%%%%%%%%%%%%%%%%%%%%%%%%%%%%%%%%%%%%%%

\end{document}